# Diffusion-RSCC: Diffusion Probabilistic Model for Change Captioning in Remote Sensing Images

*Xiaofei Yu, Yitong Li and Jie Ma*

*Abstract*—Remote sensing image change captioning (RSICC) aims at generating human-like language to describe the semantic changes between bi-temporal remote sensing image pairs. It provides valuable insights into environmental dynamics and land management. Unlike conventional change captioning task, RSICC involves not only retrieving relevant information across different modalities and generating fluent captions, but also mitigating the impact of pixel-level differences on terrain change localization. The pixel problem due to long time span decreases the accuracy of generated caption. Inspired by the remarkable generative power of diffusion model, we propose a probabilistic diffusion model for RSICC to solve the aforementioned problems. In training process, we construct a noise predictor conditioned on cross modal features to learn the distribution from the real caption distribution to the standard Gaussian distribution under the Markov chain. Meanwhile, a cross-mode fusion and a stacking self-attention module are designed for noise predictor in the reverse process. In testing phase, the well-trained noise predictor helps to estimate the mean value of the distribution and generate change captions step by step. Extensive experiments on the LEVIR-CC dataset demonstrate the effectiveness of our Diffusion-RSCC and its individual components. The quantitative results showcase superior performance over existing methods across both traditional and newly augmented metrics. The code and materials will be available online at https://github.com/Fay-Y/Diffusion-RSCC

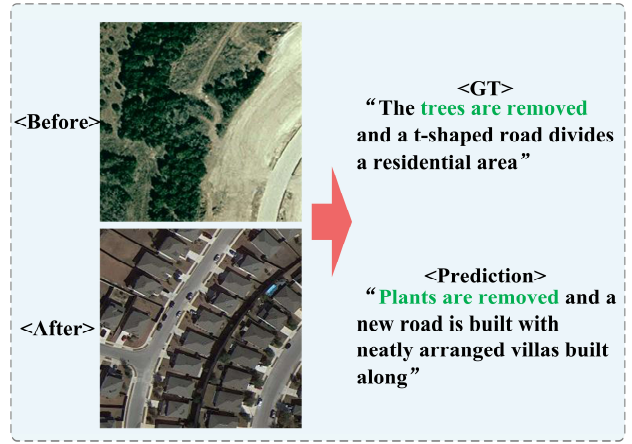

Fig. 1. An example of change captioning in Diffusion-RSCC and the comparison between model prediction and the ground truth captions.

## I. INTRODUCTION

In recent years, tasks related to multi-modal data [1-3] processing in the field of remote sensing have attracted widespread attention. Change captioning (CC) [4-6] for remote sensing (RS) images has gradually emerged as a popular theme. Derived from remote sensing change detection (RSCD) [7, 8] and image captioning (RSIC) [9], Remote sensing image change captioning (RSICC) aims at generating human-like language sentences to describe the semantic change between bi-temporal RS image pairs. It can provide accurate and insightful change captions for valuable perspectives into the dynamic nature of land cover [10, 11].

Change captioning tasks early spring up in surveillance video datasets and diagnostic datasets other than remote sensing images. The pioneer work with well-aligned image pairs in surveillance video datasets was firstly proposed in 2018 by Jhamtani et al. [12]. The initial effort established the encoder-decoder structure as the paradigm for subsequent researches. Accordingly, plenty of researchers have explored CC task and proposed various solutions to address realistic challenges, including the presence of distractors and aligning image features with textual semantics. Par et al. [13] proposed the DUDA model, a dual dynamic attention mechanism that compensated the influence of illumination variations. They further contributed by building the CLEVR-Change diagnostic dataset which contains both semantic and non-semantic changes. Qiu et al. [14, 15] extended the CC task to 3D scene, using multi-view observations. Tu et al. [16] provided a viewpoint-adaptive representation disentanglement(VARD) model, and the VARD model can deduce the changed content after removing the common properties. Some other works emphasize on the alignment of multimodality inputs. Tu et al. [17] made an advance by introducing the part-of-speech (POS) of words for switching visual information. Their attention-based visual switch (AVS) can decide when to use visual information during caption generation. Hosseinzadeh et al. [18] came up with an auxiliary task learning scheme from composed query image retrieval. The auxiliary task forced the model to pick a desired image according to captions and helped the primary task. AK et al. [19] set a text-based image manipulation (TIM) module to map the images and captions via reinforcement learning. TIM worked as a reward term to enhance caption generation.

This work was supported in part by the National Natural Science Foundation of China under Grant 62101052 and Fundamental Research Funds for the Central Universities (No. 20200D012)*(Corresponding author: Jie Ma.)*

Xiaofei Yu and Jie Ma are with the School of Information Science and Technology, Beijing Foreign Studies University, Beijing 100089, China (e-mail: majie_sist@bfsu.edu.cn).
Yitong Li is with the State Key Laboratory of Information Engineering in Surveying, Mapping and Remote Sensing, Wuhan University, Wuhan 430079, China.

Chouaf et al. [20] initiated change captioning in remote sensing, while Liu et al. [6] introduced the LEVIR-CC dataset, consisting of 10077 pairs of bi-temporal RS images and 50385 sentences describing changes. Since its release, this dataset has become a vital resource for RSICC, enabling studies to address the pixel-difference challenge in RS datasets with extensive temporal spans [6]. Different from conventional CC tasks, RSICC involves not only retrieving relevant information across different modalities [21] but also mitigating the impact of pixel-level [22] differences on terrain change localization. A primary challenge in RSICC lies in distinguishing semantic changes from spurious changes, potentially compromising caption accuracy. Liu et al. [5] addressed the challenge by decoupling the task into two stages: change detection and change captioning. By improving change localization accuracy through the change detection phase, models were able to enhance the overall precision. Furthermore, RSICC faces a dearth of high-quality datasets, as images are derived from real satellite imagery [23], and textual annotations require manual effort [6]. Limited image-text pairs pose a significant challenge for learning an appropriate mapping from changes to natural language. Additionally, auto-regressive decoders [8, 13] fail to escape the fixed structures and style of sample sentences, restricting their ability to generate flexible and authentic captions that truly align with visual features. To enhance the change captioning algorithms performance and address data insufficiency, the study must settle two crucial issues: 1) efficiently aligning the distributions of multi-temporal and multi-modal data to capture changes across time and modalities. 2) modifying the caption decoder to foster more flexible and accurate captions that are consistently aligned with the visual content.

Recently, denoising diffusion probabilistic models (DDPM) [24] have shown incredible capabilities as generative models, which are able to recreate the true sample from a Gaussian noise input. They have showcased remarkable success in both continuous visual data [24-26] and discrete language data domains [27-29], exhibiting the capability to produce high-quality samples. These models facilitate deep understanding of pixel connections, words connections and image-text connection in multi-modal tasks[30-32]. Specifically, diffusion models are a class of likelihood-based models, where a Markov chain of diffusion steps is designed to slowly add random noise to data and then learn to reverse the diffusion process to construct desired data samples from the noise. Given the inherent randomness, diffusion models are capable of capturing the diversity and uncertainty in data during the training process. This randomness not only assists in reducing overfitting for generalization capabilities, but also enables them to better accommodate the complexity and variability of real-world data. Considering the desirable properties of diffusion models, such as a stationary training objective, and easy scalability, we aim to bring the benefits to the field of change captioning and overcome the abovementioned challenges.

In this paper, we propose a novel probabilistic diffusion model to capture semantic change in bi-temporal remote sensing images. The proposed framework aims to learn the mapping from the standard Gaussian distribution to the distribution of real captions through condition denoiser. As for testing phase, model can incrementally denoise the standard Gaussian noise into vectors corresponding to words. The gradual denoising steps help produce hierarchical continuous latent representations and generate change caption under image control. Specifically, the proposed condition denoiser is composed of five main blocks:1) pretrained CNN to extract features from the bi-temporal image pairs; 2) time embedding block and 3) position embedding block to incorporate the temporal and spatial information; 4) cross-mode fusion (CMA) module to integrate the visual features and embedded sentences with spatiotemporal precision; and 5) staking self-attention (SSA) module to iteratively refine the mutual attention score between cross-modal data for conditional mean value estimation. The condition denoiser, especially the CMA and SSA module, helps the diffusion model to present semantic changes with well-aligned input mapping. Based on the learning principle and data integration across modalities, RSCC-Diffusion can comprehend contextual information and generate semantically accurate and fluent descriptions. Through this work, we significantly advance the field of remote sensing image change captioning with a non-autoregressive framework.

To sum up, our contributions are two-fold:

1) ***The enhancement of factual correctness and flexibility:*** The proposed diffusion-based generative model can learn the cross-modal data distribution between the image pairs and change captions, and the randomness in diffusion models helps reduce overfitting and bring better complexity and variability to generated captions.
2) ***The construction of cross-modal mapping:*** The proposed cross-mode fusion and staking self-attention in condition denoiser effectively establish mappings between visual features and textual sequences, facilitating conditional mean value estimation during the reverse process. Ablation studies demonstrate the effectiveness of these modules.

## II. RELATED WORK

### A. Diffusion Process

Diffusion probabilistic models emerge as formidable contenders alongside convolutional neural networks (CNNs) and attention in computer vision and natural language processing. They have showcased remarkable generative power in artificial intelligence generated content.

Diffusion model works as a parameterized Markov chain trained using variational inference to produce samples matching the data after finite time. It consist of two-phased process: in forward diffusion process, model gradually adds noise to the data by sampling until signal is destroyed while in the reverse process, model uses a noise predictor, for example U-Net [33], to denoise the destroyed data from the forward process to achieve complete representation restoration.

The generative capabilities of diffusion models have been leveraged for robust feature extraction and representation. Since Denoising Diffusion Probabilistic Models (DDPM) [24] firstly introduced diffusion into computer vision, diffusion models find diverse applications in visual tasks such as video generation [34-36], image editing [37, 38], and image super-resolution [25, 39, 40]. In remote sensing, IDF-CR [41] consisted of a pixel space cloud removal (Pixel-CR) module and a latent space iterative noise diffusion (IND) network and exhibits strong generative capabilities to achieve component divide-and-conquer (CDC) cloud removal. EDiffSR [42] with a novel noise predictor EANet can restore visual-pleasant images on both simulated and real-world remote sensing images.

Diffusion models have been extended to numerous fields. ControlNet [43] introduced multiple auxiliary condition paths, while Stable Diffusion ensured a stable diffusion process by projecting diffusion and sampling into the latent space. Diffusion-LM[28] introduced the diffusion model into natural language processing and used the BERTEncoder [44] as the noise predictor. It successfully transmitted the diffusion models from continuous image data to discrete words and showed the potential for the application in multi-modal tasks. Glide [45] with a U-Net noise predictor explored diffusion models for the problem of text-conditional image synthesis and it can be finetuned to perform image inpainting, enabling powerful text-driven image editing. SCD-Net [31] coupled the continuous diffusion process with a new guided transformer-based sequence learning in image captioning.

However, the utilization of diffusion models in remote sensing cross-modal tasks remains largely unexplored. Inspired by these outstanding efforts, we leverage diffusion models with a new attention-based noise predictor for more accurate advanced semantics and richer fine-grained details in remote sensing image change captioning.

*B. Change Detection in Remote Sensing*

Remote sensing change detection (RSCD) has evolved from traditional approaches to advanced deep learning methods. These traditional techniques laid the groundwork for current remote sensing change detection methodologies.

Change vector analysis (CVA) [46] method calculated change intensity and direction between pixels to distinguish change regions using thresholds. Principle component analysis (PCA) [47] in change detection helped figure out variance in multi-temporal images, while Multivariate Alteration Detection (MAD) [48] method was applied to detect the semantic changes of corresponding multitemporal image scenes. The advanced method Li-Strahler geometric-optical model [49] can combine with a scaling-based approach to detect structural changes.

With the advent of deep learning, remote sensing change detection has seen substantial improvements. Deep learning models, including convolutional neural networks (CNNs) and Transformers, have enhanced the performance of change detection by leveraging superior multi-temporal feature learning capabilities. DDCNN [50] with a dense attention method used the spatial context information to capture the changed features of ground objects. The PA-Former [51] can combine CNN-based feature extraction with Transformer encoder and decode for refined change discrimination. Diffusion models also play an important role in RSCD. GAD algorithm [52] fitted semantic segmentation predictions to the input images for weakly supervised change detection. SwiMDiff [53] integrated contrastive learning with a pixel-level diffusion model and has been rigorously tested on change detection.

All these methods provide valuable insights for the utilization of diffusion in remote sensing image change captioning.

*C. Change Captioning*

Change captioning is an emerging task in the vision-language community that extends beyond conventional image or video captioning by describing the differences between bi-temporal images. Unlike traditional captioning, which focuses on a single image or video frame, change captioning must handle the added complexity of identifying and articulating the changes between image pairs.

Early work in change captioning by Jhamtani et al. [12] focused on well-aligned image pairs, simplifying the challenge by ensuring minimal viewpoint variations. To address the practical issue of viewpoint changes in dynamic environments, Park et al. [13] introduced the CLEVER-Change dataset, incorporating both semantic and pseudo changes due to viewpoint shifts. The proposed DUDA by Park et al. localized and described changes by direct subtraction of bi-temporal image pairs. SRDRL [17] introduced the part-of-speech (POS) of words for switching visual information. CC-Full [19] combined reinforcement learning with text-based image manipulation TIM to refine change localization and captioning.

Other works focus on explicitly modeling the features of change which contributes to the alignment of multimodality input. VAM [54] employed a viewpoint-agnostic matching encoder to summarize and remove common properties between images to infer changes. VACC [55] predicted change features and refined sentences using a cycle consistency module. $R^3$Net [56] utilized a representation reconstruction network and a syntactic skeleton predictor to enhance semantic interactions between change localization and caption generation. MCCFormers-S and MCCFormers-D [57] leveraged a Transformers encoder-decoder structure to model interactions between images implicitly and caption generation. Recent advancements have further refined these approaches. VARD [16] disentangled viewpoint and pseudo changes for more accurate change localization. These diverse approaches make rapid progress to tackle the challenges posed by viewpoint changes and pseudo changes in change captioning.

However, the remote sensing image change captioning task faces more challenges of limited datasets and pseudo pixel-difference distractors. RSICCformer [6] with the proposed LEVIR-CC dataset set the benchmark for RSICC. PromptCC [5] addressed the primary challenge by distinguishing visual semantics from spurious changes and decoupling the task into change detection and captioning stages for GPT prompt. To enhance algorithm performance, works in remote sensing change captioning should address data insufficiency by efficiently aligning the distributions of multi-temporal and multi-modal data and modifying the caption decoder to generate more flexible and accurate captions. Inspired by previous work, we introduce the diffusion models to achieve better cross-modal alignment in RSICC.

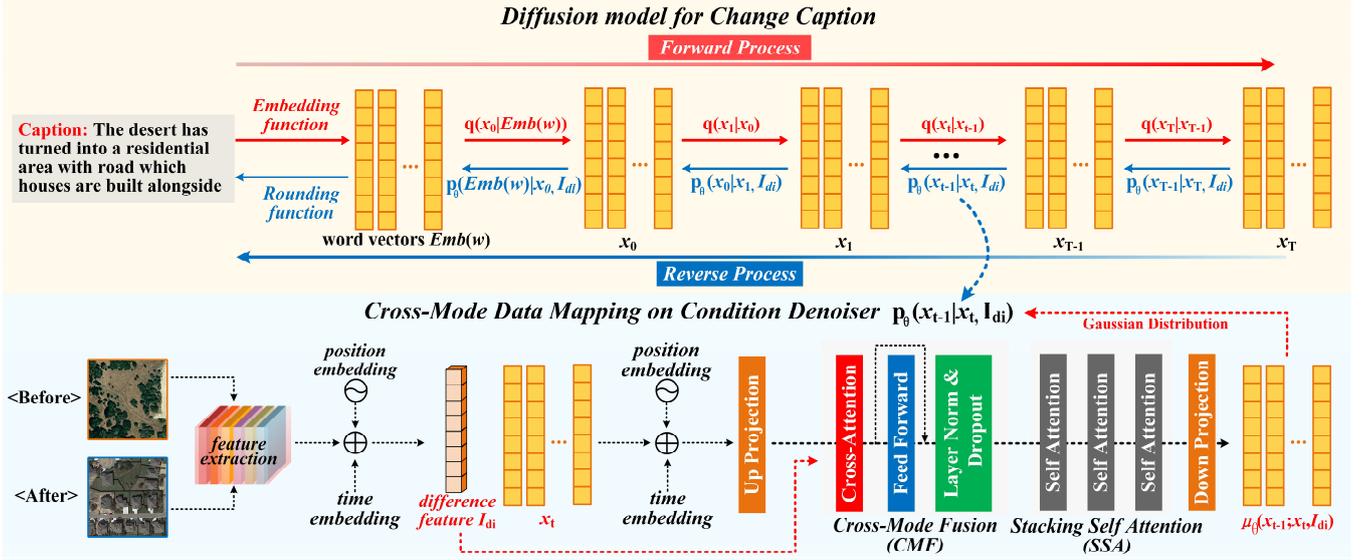

Fig. 2. The flowchart of our proposal. The diffusion framework consists of two processes: forward process to add noise to data until corruption and reverse steps to denoise with the condition denoiser. The condition denoiser consists of 1) CNN-based feature extractor, 2) time embedding block, 3) position embedding block, 4) cross-mode fusion (CMF) module and 5) staking self-attention (SSA) module.

## III. METHODOLOGY

In this section, we propose a novel diffusion-based generative model for RSICC, as illustrated in Fig.2. The proposed denoising diffusion probabilistic models is composed of three main parts: 1) the forward diffusion process, 2) the reverse denoising process and 3) the construction of condition denoiser for CC task. The purpose of the forward diffusion process is to construct a mapping from the distribution of caption to standard Gaussian distribution, while the reverse process aims to denoise the prior standard Gaussian distribution back to the real data distribution using denoiser. It's worth to mention that the construction of denoiser plays a vital role in our model, which establishes the mapping from image to language and decides the accuracy of the captions directly.

In the following part, we firstly introduce the forward and backward step of the diffusion models in section A and B. Then, we will present the network architecture of the denoiser in section C. The implementation details in training and testing phase are presented in section D.

### A. The Forward Diffusion Process

Diffusion models have shown great performance in continuous data domains, such as image [24] and video generation [36]. As for discrete language data, the main challenge is to learn the continuous latent representations of discrete words. Inspired by previous work proposed by Li et al. [28], we firstly construct a learnable embedding module in the forward diffusion process, that establishes the mapping from discrete word to the continuous representation space.

Change captioning task aims to sample $w$ from a trainable language model $p(w | I_{bef}, I_{aft})$, where $w = [w_1, w_2, \cdots, w_n]$ is the sentence of discrete word with length $n$, $I_{bef}$ and $I_{aft}$ are the remote sensing image pairs obtained over the same area with long time span. We define the word embedding function $Emb(w_i): \mathcal{R} \to \mathcal{R}^d$ to transit discrete words to continuous data distribution. Then the embedding of the sentence $w$ can be computed as:

$$Emb(w) = [\ Emb(w_1), Emb(w_2), \cdots, Emb(w_n)\ ]. \quad (1)$$

In forward process, noise is added following a Markov transition [28] from embedding representation $Emb(w)$ to $x_0$, following $q_\phi(x_0 | w) = \mathcal{N}(Emb(w), (1-\alpha_0)\mathbf{I})$, where $\alpha_0 \in \mathcal{R}$ is a hyper-parameter, $\phi$ is trainable parameters and $\mathbf{I} \in \mathcal{R}^{n \times d}$ is an identity matrix. Then according to continuous diffusion step [58], the forward process can be formulated as:

$$q(x_t | x_{t-1}) := \mathcal{N}(x_t; \sqrt{\alpha_t} x_{t-1}, (1-\alpha_t)\mathbf{I}), \quad (2)$$

Where $\alpha_t$ is a hyper-parameter. Different from continuous situation, by composition of forward process from $w$ to $x_T$, the joint distribution under $w$ following the Markov chain can be computed as:

$$q(x_{0:T} | w) = q(x_0 | w) \prod_{t=1}^{T} q(x_t | x_{t-1}), \quad (3)$$

Inspired by [58], suppose that we have $2T$ random noise variables $\{\epsilon_t^*, \epsilon_t\}_{t=1}^{T}$ independent identically distributed sampling from $\mathcal{N}(0, \mathbf{I})$, we can sample any arbitrary sample $x_t$ by reparameterization trick:

$$\begin{aligned}
x_t &= \sqrt{\alpha_t} x_{t-1} + \sqrt{1-\alpha_t}\, \epsilon_{t-1}^* \\
&= \sqrt{\alpha_t}\left(\sqrt{\alpha_{t-1}} x_{t-2} + \sqrt{1-\alpha_{t-1}}\, \epsilon_{t-2}^*\right) + \sqrt{1-\alpha_t}\, \epsilon_{t-1}^* \\
&= \sqrt{\alpha_t \alpha_{t-1}} x_{t-2} + \sqrt{\alpha_t - \alpha_t \alpha_{t-1}}\, \epsilon_{t-2}^* + \sqrt{1-\alpha_t}\, \epsilon_{t-1}^* \\
&= \sqrt{\alpha_t \alpha_{t-1}} x_{t-2} + \sqrt{\sqrt{\alpha_t - \alpha_t \alpha_{t-1}}^2 + \sqrt{1-\alpha_t}^2}\, \epsilon_{t-2} \\
&= \cdots \\
&= \sqrt{\prod_{i=1}^{t} \alpha_i}\, x_0 + \sqrt{1-\prod_{i=1}^{t}\alpha_i}\, \epsilon_0 \\
&\sim \mathcal{N}(x_t; \sqrt{\bar{\alpha}_t} x_0, (1-\bar{\alpha}_t)\mathbf{I})
\end{aligned} \quad (4)$$

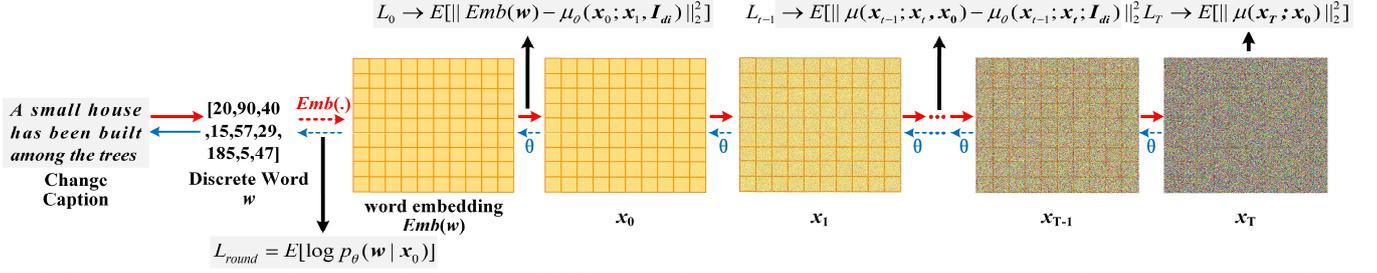

Fig.3. The training process of the proposed diffusion model. The forward noising process is in red and the reverse denoising process is in blue. The black arrows point out the four components of the training objective.

where $\bar{\alpha}_t = \prod_{i=1}^{t} \alpha_i$, $\alpha_i \in (0,1)$. The parameters $\alpha_i$ and $\alpha_0$ evolve over time according to a fixed or learnable schedule structured such that the distribution of the final latent $x_T$ is a standard Gaussian [24]. Given the forms of $x_t$, a random variable $\epsilon \sim \mathcal{N}(0,\mathbf{I})$ and time step $t$, we can obtain the latent representation in only one step.

It has been demonstrated that [58], knowing the forms of both $q(x_t|x_0)$ and $q(x_{t-1}|x_0)$, $q(x_{t-1}|x_t,x_0)$ can be calculated as a Gaussian distribution $\mathcal{N}(\mu(x_{t-1};x_t,x_0),\Sigma(t))$, where:

$$\mu(x_{t-1};x_t,x_0) = \frac{\sqrt{\alpha_t}(1-\bar{\alpha}_{t-1})x_t + \sqrt{\bar{\alpha}_{t-1}}(1-\alpha_t)x_0}{1-\bar{\alpha}_t}, \quad (5)$$

$$\Sigma(t) = \frac{(1-\alpha_t)(1-\bar{\alpha}_{t-1})}{1-\bar{\alpha}_t}\mathbf{I}. \quad (6)$$

### B. The Reverse Diffusion Process

Firstly, we establish a rounding step with trainable parameters to get the discrete words from the continuous embeddings:

$$p_\theta(\mathbf{w}|x_0) = \prod_{i=1}^{n} p_\theta(w_i|x_{0,i}), \quad (7)$$

where $x_0 = [x_{0,1},...,x_{0,i},...x_{0,n}]$ and $p_\theta(w_i|x_{0,i})$ is a linear projection layer followed by SoftMax activation function.

Assuming that $p_\theta(x_{t-1}|x_t,I_{bef},I_{aft})$ is a learnable distribution that can estimate latent representation of previous state $x_{t-1}$ when given $x_t$ and the bi-temporal image pair $I_{bef}$ and $I_{aft}$. Following [28], the training objective aims to maximize the Evidence Lower Bound (ELBO) of log likelihood of $p(\mathbf{w}|I_{bef},I_{aft})$, which can be computed as:

$$\log[p(\mathbf{w}|I_{bef},I_{aft})] \geq E_{q_\phi(x_{0:T}|\mathbf{w})}[\log\frac{q(x_T|x_0)}{p_\theta(x_T)}$$
$$+ \sum_{t=2}^{T}\log\frac{q(x_{t-1}|x_0,x_t)}{p_\theta(x_{t-1}|x_t,I_{bef},I_{aft})}$$
$$+ \log\frac{q_\phi(x_0|\mathbf{w})}{p_\theta(x_0|x_1,I_{bef},I_{aft})} \quad (8)$$
$$- \log p_\theta(\mathbf{w}|x_0)]$$

We can rewrite each term in (8) as:

$$L = L_T + \sum_{t=2}^{T} L_{t-1} + L_0 - L_{round}. \quad (9)$$

And the four terms can take the following form:

$$L_T = E_{q_\phi(x_{0:T}|\mathbf{w})}[\log\frac{q(x_T|x_0)}{p_\theta(x_T)}], \quad (10)$$
$$= D_{KL}(q(x_T|x_0) \| p_\theta(x_T))$$

where $q(x_T|x_0)$ and $p_\theta(x_T)$ are both Gaussian distributions. Considering that $p_\theta(x_T)$ is $\mathcal{N}(x_T;0,\mathbf{I})$ when $T$ is sufficiently large, the KL Divergence can be simplified as follows [28]:

$$L_T \to E[\|\mu(x_T;x_0) - 0\|_2^2] = E[\|\mu(x_T;x_0)\|_2^2], \quad (11)$$

Similarly, we have:

$$L_{t-1} \to E_{q_\phi(x_{0:T}|\mathbf{w})}[\log\frac{q(x_{t-1}|x_0,x_t)}{p_\theta(x_{t-1}|x_t,I_{bef},I_{aft})}] \quad (12)$$
$$= E[\|\mu(x_{t-1};x_t,x_0) - \mu_\theta(x_{t-1};x_t,I_{bef},I_{aft})\|_2^2]$$

$$L_0 \to E_{q_\phi(x_{0:T}|\mathbf{w})}[\log\frac{q_\phi(x_0|\mathbf{w})}{p_\theta(x_0|x_1,I_{bef},I_{aft})}] \quad (13)$$
$$= E[\|Emb(\mathbf{w}) - \mu_\theta(x_0;x_1,I_{bef},I_{aft})\|_2^2]$$

where $L_{t-1}$ and $L_0$ are converted into the mean square error between the forward mean value and backward mean value at corresponding step $t$.

Then, the loss of rounding step $L_{round}$ can be recalculated as the cross-entropy form:

$$L_{round} = E_{q_\phi(x_{0:T}|\mathbf{w})}[\log p_\theta(\mathbf{w}|x_0)], \quad (14)$$

where the parameters $\phi$ of embedding function $Emb(\cdot)$ and rounding function $p_\theta(\mathbf{w}|x_0)$ can be trained in $L_0$, $L_{round}$, $L_T$.

As shown in (12), assuming that $p_\theta(x_{t-1}|x_t,I_{bef},I_{aft})$ is Gaussian distribution, the form of $p_\theta(x_{t-1}|x_t,I_{bef},I_{aft})$ can be:

$$\mathcal{N}(\mu_\theta(x_{t-1};x_t,I_{bef},I_{aft});\Sigma(t)), \quad (15)$$

where the denoised mean value can be written as:

$$\mu_\theta(x_{t-1};x_t,I_{bef},I_{aft}) = \frac{\sqrt{\alpha_t}(1-\bar{\alpha}_{t-1})x_t + \sqrt{\bar{\alpha}_{t-1}}(1-\alpha_t)f_\theta(x_t,t,I_{bef},I_{aft})}{1-\bar{\alpha}_t}, \quad (16)$$

and the variance can be computed as (6). Thus, the key step of the reverse diffusion process is how to estimate the mean value of the distribution when given the time step $t$, the latent representation $x_t$ and the feature of bi-temporal image pair $I_{bef}, I_{aft}$. In the subsequent part, we construct a condition denoiser $f_\theta(x_t,t,I_{bef},I_{aft})$, that can help to successfully predict the denoised mean value of $p_\theta(x_{t-1}|x_t,I_{bef},I_{aft})$.

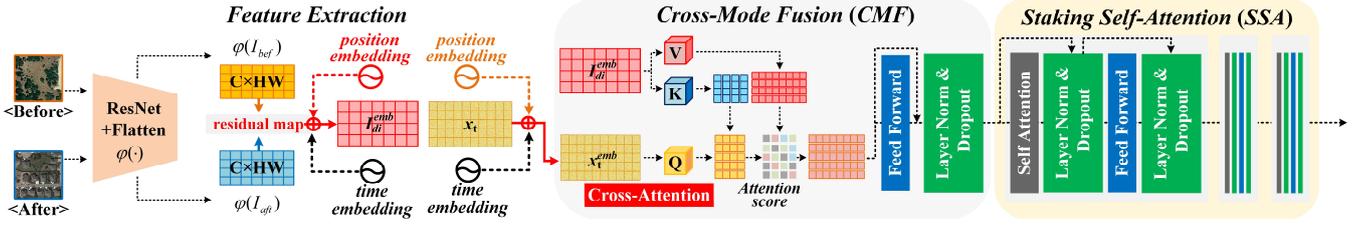

Fig. 4. Network architecture of the proposed condition denoiser:1) feature extraction, 2) position embedding,3) time embedding,4) cross-mode fusion (CMF) and 5) staking self-attention (SSA).

*C. Cross-Mode Data Mapping on Condition Denoiser*

In this part, we construct a condition denoiser network based on cross-modal attention mechanism for the reverse diffusion process. The network architecture is shown in Fig. 4, the detailed network architecture is as follows.

**1) Feature extraction:** Inspired by the previous works[4-6, 13], we encode each "before-after" image pair $I_{bef}, I_{aft}$ with ResNet101 [59] to extract the high level feature representations $\varphi(I_{bef})$, $\varphi(I_{aft}) \in \mathbb{R}^{C \times H \times W}$, where $C$, $H$, $W$ indicate the number of channels, height and width. And then the extracted features are flattened to $C \times HW$.

To capture the distinction of the bi-temporal image pair, we obtain the residual map $I_{di}$ of the flattened features by:

$$I_{di} = \varphi(I_{bef}) - \varphi(I_{aft}). \quad (17)$$

Then the residual maps are fed to the condition denoiser $f_\theta(x_t, t, I_{bef}, I_{aft}) = f_\theta(x_t, t, I_{di})$ and help to compute the mean value of $p_\theta(x_{t-1} | x_t, I_{bef}, I_{aft}) = p_\theta(x_{t-1} | x_t, I_{di})$ in reverse process.

**2) Position embedding:** As shown in Fig. 4, the position embedding technique is introduced in both feature extraction $I_{di}$ and latent representation of $x_t$. The main reason is that, we propose an attention-based network for denoising step in the following part. The sine and cosine functions of different frequencies are used for position embedding as (18):

$$PE_{(pos, 2i_P)} = sin(pos / 10000^{2i_P / d_{model}})$$
$$PE_{(pos, 2i_P+1)} = cos(pos / 10000^{2i_P / d_{model}}), \quad (18)$$

where $pos$ is the position of current token and $i_P$ is the dimension. Each dimension of the positional encoding corresponds to a sinusoid [60].

**3) Time embedding:** As shown in Fig. 4, time embedding is introduced in the latent representation of $x_t$ since the proposed diffusion-based framework. Similar to position embedding [61], time embedding are also devised on sine and cosine functions as shown below:

$$TE_{(t, 2i_t)} = sin(t / 10000^{2i_t / d_{model}})$$
$$TE_{(t, 2i_t+1)} = cos(t / 10000^{2i_t / d_{model}}), \quad (19)$$

where $t$ is the timestep and $i_t$ is the dimension.

**4) Cross-Mode Fusion (CMF):** CMF combines the cross attention and multi-head attention to compute the mutual weights between semantic image feature and textual sequence for well-aligned mapping. As shown in Fig. 4, CMF takes $x_t^{emb}, I_{di}^{emb}$ as the input, where the embedded representation $x_t^{emb}$ and $I_{di}^{emb}$ are from $x_t$ and $I_{di}$ concatenated with position embedding $PE$ and time embedding $TE$ as (20):

$$\begin{aligned} x_t^{emb} &= x_t + PE + TE \\ I_{di}^{emb} &= I_{di} + PE + TE \end{aligned}. \quad (20)$$

As for the cross multi-head attention module, the query is from $x_t^{emb}$, while key and value come from the change feature $I_{di}^{emb}$. And the cross multi-head attention $CMHA(\cdot,\cdot)$ in CMF module can be formulated as:

$$CMHA(x_t^{emb}, I_{di}^{emb}) = Concat(head_1, \ldots, head_h), \quad (21)$$
$$head_j = MultiheadAttention(x_t^{emb} W_{hj}^Q, I_{di}^{emb} W_{hj}^K, I_{di}^{emb} W_{hj}^V), \quad (22)$$

where $W_{hj}^Q \in \mathbb{R}^{d_q \times d_{model}}$, $W_{hj}^K \in \mathbb{R}^{d_k \times d_{model}}$, $W_{hj}^V \in \mathbb{R}^{d_v \times d_{model}}$ are the parameter matrices of the $j$-th head, $d_{model}$ is the hidden size.

To increase the nonlinearity of the network, additive attention computes a compatibility function using a feed-forward network which consists of two linear layers FC with a ReLU activation function. It can be formulated as follows:

$$x_t^{proj} = FC(ReLU(FC(CMHA(x_t^{emb}, I_{di}^{emb})))), \quad (23)$$
$$x_t^{fus} = LN(x_t^C + Dropout(x_t^{proj})), \quad (24)$$

where LN denotes a Layer Norm and $x_t^C$ is the output of $CMHA(\cdot,\cdot)$. Then, the fine-grained mapping output $x_t^{fus}$ of CMF are fed to the following staking self-attention module.

**5) Staking Self-Attention (SSA):** Self-attention can recompute the representation of input $x_t^{fus}$, relating different positions of a single sequence. Through the staking self-attention, the condition denoiser helps reduce $x_t^{fus}$ noise and predict a new mean value with structural coherence. The query, key and value all derive from the CMF output $x_t^{fus}$.

$$x_t^l = Self\text{-}attention_l(x_t^{fus} W_l^Q, x_t^{fus} W_l^K, x_t^{fus} W_l^V), \quad (25)$$

where $W_l^Q$, $W_l^k$, $W_l^V \in \mathbb{R}^{d \times d_{model}}$ are trainable matrices of the $l$-th self-attention layer.

After the self-attention layer (25), the $x_t^l$ is passed through a complementary attention block which takes following terms:

$$x_t^{res} = LN(Dropout(FC(x_t^l)) + x_t^l), \quad (26)$$
$$x_t^{act} = GeLU(FC(x_t^{res})), \quad (27)$$
$$f_\theta(x_t, t, I_{di}) = LN(x_t^{act} + Dropout(FC(x_t^{act}))). \quad (28)$$

According to (16), model can use the attention-based condition denoiser to reconstruct the mean value $\mu_\theta(x_{t-1}; x_t, I_{bef}, I_{aft})$ of the reverse process at step $t$ and force the model to denoise for natural caption generation $w$ under the image feature $I_{di}$.

## D. Training and Testing

In training phase (**Algorithm 1**), we successively optimize the training objective for the condition denoiser and the rounding function. In testing phase (**Algorithm 2**), we estimate the distribution of $p_\theta(x_{t-1}|x_t, I_{bef}, I_{aft}) = p_\theta(x_{t-1}|x_t, I_{di})$ through the condition denoiser and predict high-quality change captions $w$ on image pair $I_{bef}$ and $I_{aft}$.

---

**Algorithm 1 Training phase**

**Input:** Total diffusion timestep $T$, caption $w$, image pair $I_{bef}$ and $I_{aft}$, hyper-parameters $\alpha_0, \alpha_1, \ldots \alpha_T$, position embedding $PE$, time embedding $TE$.

1: **Initialize:** ResNet101 network with flatten $\varphi(\cdot)$, embedding function $Emb(\cdot)$, condition denoiser $f_\theta(\cdot)$, rounding function $p_\theta(\cdot)$.

2: **repeat:**

3:  Compute the image feature: $I_{di} = \varphi(I_{bef}) - \varphi(I_{aft})$.

4:  Select randomly $t \sim Uniform(\{1,2,\ldots,T\}, \epsilon \sim \mathcal{N}(0,\mathbf{I})$.

5:  Compute the start point $x_0 = Emb(w) + \sqrt{1-\alpha_0}\epsilon$.

6:  Compute current step at $t$: $x_t = \sqrt{\bar{\alpha}_t}x_0 + \sqrt{1-\bar{\alpha}_t}\epsilon$.

7:  Compute the condition denoiser result: $f_\theta(x_t, t, I_{di})$.

8:  Predict denoised mean value $\mu_\theta(x_{t-1}; x_t, I_{di})$ according to (16).

9:  According to (9-14), compute the training objective:

9:  **If** t=0:
$$L = (\sqrt{\bar{\alpha}_T}x_0)^2 + (Emb(w) - \mu_\theta(x_t,t,I_{di}))^2 - \log(p_\theta(w|x_0)).$$

9:  **Else**:
$$L = (\sqrt{\bar{\alpha}_T}x_0)^2 + (x_0 - f_\theta(x_t,t,I_{di}))^2 - \log(p_\theta(w|x_0)).$$

10:  Optimize $\theta$ by minimizing $L$.

11: **Until** converged

12: **return** $\theta$

---

**Algorithm 2 Testing phase**

**Input:** Total diffusion timesteps $T$, image pair $I_{bef}$, $I_{aft}$, hyper parameters $\alpha_0,\ldots,\alpha_T$, position embedding $PE$, time embedding $TE$.

**Load:** ResNet101 with flatten $\varphi(\cdot)$, condition denoiser $f_\theta(\cdot)$, rounding function $p_\theta(\cdot)$

1: Compute the image feature: $I_{di} = (\varphi(I_{bef}) - \varphi(I_{aft}))$

2: Initialize $x_T \sim N(0,\mathbf{I})$

3: **for** $t = T, T-1, \ldots, 1$ **do**

4:  Sample $\epsilon_t \sim N(0,\mathbf{I})$

5:  Sample $x_{t-1}$ from denoised distribution $\mathcal{N}(\mu_\theta(x_{t-1}; x_t, I_{di}), \Sigma(t))$:
$$x_{t-1} = \frac{\sqrt{\alpha_t}(1-\bar{\alpha}_{t-1})x_t + \sqrt{\bar{\alpha}_{t-1}}(1-\alpha_t)f_\theta(x_t,t,I_{di})}{1-\bar{\alpha}_t}$$
$$+ \frac{(1-\alpha_t)(1-\bar{\alpha}_{t-1})}{1-\bar{\alpha}_t}\bullet\epsilon_t$$

6: **end for**

7: Select word $w_i$ with highest probability $p_\theta(w|x_0)$ for token $x_{0,i}$:
$$p_\theta(w|x_0) = \prod_i^n p_\theta(w_i|x_{0,i})$$

8: **return** $w = [w_0,\ldots,w_i,\ldots,w_n]$ as change caption prediction

---

## IV. EXPERIMENTS

In this section, we describe the experimental results of our model and previous state-of-the-art methods through several experimental setups on LEVIR-CC dataset. We also conduct ablation experiments on different sub-structures, including image feature extraction, mapping of image features and caption vectors and self-attention blocks to demonstrate the effectiveness of each module.

### A. LEVIR-CC Dataset

LEVIR-CC is a large-scale change captioning dataset proposed by Liu et al [6]. It contains 10077 image pairs (5038 changed image pairs and 5039 unchanged image pairs) from 20 different regions in Texas, USA. The images cover a time span ranging from 5 to 15 years and are sized at 256×256 pixels with a resolution of 0.5m/pixel. Within this dataset, each changed image pair exhibits one to two instances of alternation with 5 kinds of ground surface changes including buildings, roads, parking lots, vegetation, and water. Each image pair has five captions, 50385 descriptive captions in total. The maximum caption sentence length is 39 words, with an average of 7.99 words. For training and testing, the dataset is partitioned into training, validation, and testing sets with 6,815, 1,333, and 1,929 samples. Our vocabulary dictionary comprises the words that appear in the annotated captions, totaling 998 unique words. Notably, while previous works[5, 6] tend to remove words with a frequency below 5 to reduce data sparsity, we retain these low-frequency words to preserve the semantic integrity of the captions and ensure that the model could learn from a wider vocabulary distribution.

### B. Experimental Setup

**1) Evaluation Metrics:** The evaluation of the change captioning models is essential in assessing how well the generated sentences align with human judgments of the changes between image pairs. Traditional automatic evaluation metrics provide a measure of the accuracy of the generated captions based on annotated reference sentences. In this paper, we utilize five commonly used metrics in change captioning tasks [13, 56, 62] to evaluate the accuracy of all methods: BLEU-N (where N=4) [63], ROUGE-L[64], METEOR[65], CIDEr [66] and SPICE [67]. These metrics calculate the consistency between the predicted sentences and reference sentences. The candidates with higher scores indicate a greater degree of similarity and higher captioning accuracy. Specifically, we augment traditional metrics with additional indicators that are more suitable for evaluating captions with high variability: BARTScore[68] and MoverScore. In the augmented metrics, candidates with higher scores also indicate higher captioning consistency.

**2) Implementation Details:** We implemented all model training and evaluation on the Torch DL framework with a 24-GB NVIDIA RTX 4090 GPU. Following prior works, we use the ResNet101 network pretrained on ImageNet to extract image features. In training phase, the model is trained for 564 epochs and 2000 diffusion steps with a batch size of 32. The total number of parameters is 58,423,318. We build the vocabulary dictionary with a scale of 998 and set the dimension of word embedding to 16. We utilize the AdamW optimizer with an initial learning rate 1e-4.

TABLE I
QUANTITATIVE COMPARISON OF DIFFUSION-RSCC WITH OTHER CHANGE CAPTIONING MODELS. OUR PROPOSED MODEL OUTPERFORMS ALL BASELINES ON BLEU-4, METEOR, ROUGE-L, BARTSCORE AND MOVERSCOE ON THE LEVIR-CC DATASET. THE NUMBERS ARE IN %. THE BEST RESULTS ARE IN **BOLD**

| Method | BLEU-4 | METEOR | ROUGE-L | CIDEr | SPICE | BARTScore | MoverScore |
|---|---|---|---|---|---|---|---|
| DUDA | 57.6 | 37.7 | 71.1 | **126.6** | 29.3 | -419.6 | 30.9 |
| RSICCformer | 49.6 | 28.5 | 59.3 | 97.4 | 20.9 | -441.1 | 24.9 |
| MCCFormers-S | 46.2 | 27.22 | 58.0 | 94.3 | 20.3 | -443.6 | 23.9 |
| MCCFormers-D | 31.5 | 24.2 | 52.1 | 77.4 | 22.2 | -450.1 | 24.5 |
| VARD | 43.2 | 33.5 | 66.6 | 113.7 | **33.9** | -437.3 | 30.2 |
| Diffusion-RSCC(ours) | **60.9** | **37.8** | **71.5** | 125.6 | 26.7 | **-417.7** | **32.3** |

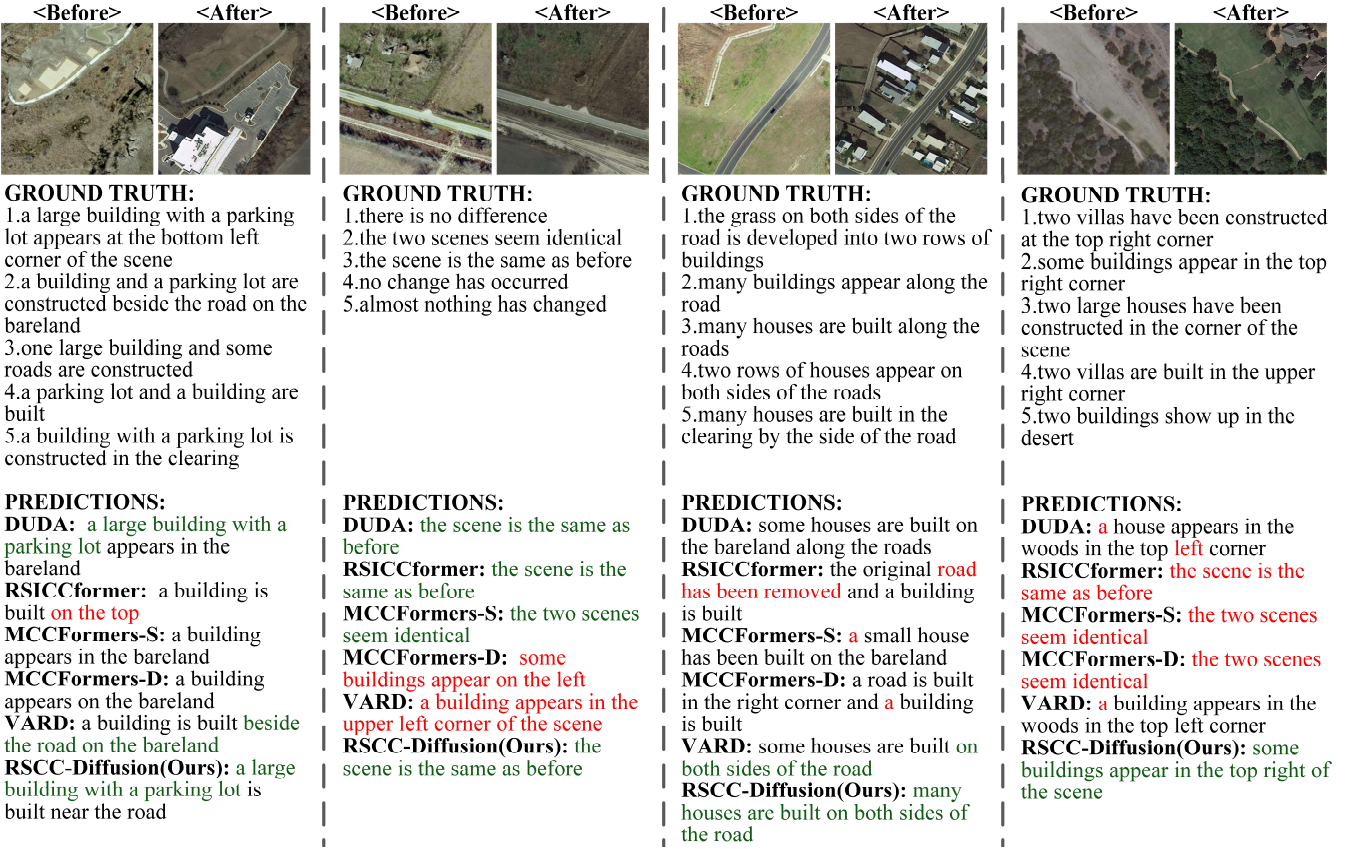

Fig. 5. Quantitative examples on the test split of LEVIR-CC dataset. We report the captions obtained by our method and all baselines along with the ground truth. Correct sequences are in green and incorrect parts of the captions are in red.

## C. Quantitative Results

We compare the proposed method Diffuison-RSCC with previous state-of-the-art methods on the LEVIR-CC dataset, including DUDA [13], MCCFormers-S [57], MCCFormers-D [57], RSICCformer [6] and VARD [16]. These methods are constructed on the encoder-decoder framework, in which the encoder extracts high-level image features and the decoder utilizes the LSTM-based or Transformer-based blocks to generate predicted captions. All the methods are described in details as follows:

1) DUDA [13]: Par et al.[13] firstly exploited the change captioning problem with data not well-aligned. There exists illumination and viewpoint change distractors in the proposed dataset CLEVR-Change by Par et al. [13] and these distractors decreases the prediction accuracy. The DUDA model consists of a dual attention module for feature extraction and mitigating the impact of distractors, and a LSTM-based dynamic speaker module for caption generation.

2) MCCFomers-S [57]: To improve the process of caption encoding, MCCFormers-S uses a Transformer caption decoder. And for the robustness to localize multiple change, Qiu et al. [57] devised two Transformer-based encoding approach to extract high-level image features. The first encoding approach MCCFormers-S directly feeds the feature from ResNet to the Transformer-based encoder.

3) MCCFomers-D [57]: MCCFormers-D uses the second image encoding approach devised by Qiu et al. [57] which sets a Siamese Transformers with cross-attention mechanism for feature extraction. Then the Transformer decoder uses obtained image features to generate captions.

4) RSICCformer [6]: RSICCformer [6] is firstly designed for remote sensing image pairs. It uses a Siamese Cross-attention module with difference feature to improve the feature discrimination capacity for changes. And multiple Transformer decoder layers are stacked to progressively utilize the image feature from ResNet block to recognize changes of interest and generate captions.

5) VARD [16]: For further mitigating the impact of distractors, VARD sets a projection-based encoder including: position-embedded representation learning, unchanged representation disentanglement and neighbor-aware position encoding. VARD [16] successfully reduces the unchanged objects for change localization. The extracted features are then fed into the LSTM or Transformer caption generator.

We design a model structure that is completely different from previous work, reducing reliance on the encoder-decoder framework. We no longer use LSTM or Transformer for caption generation, but instead rely on probabilistic diffusion mechanism. Table I summarizes the quantitative results of our model and all the baselines. Our model outperforms all tested baseline models in most metrics. This demonstrates the generalizability of our Diffusion-RSCC. Although in the metric CIDEr column, our model isn't the top performer, the difference compared to the best baseline is marginal, further validating the effectiveness of our approach.

Diffusion-RSCC model performs best on traditional metrics, showcasing the predictions' accuracy and similarity with ground-truth captions, while it also excels in the newly proposed metrics, demonstrating its capability to maintain caption consistency while enhancing text diversity.

### D. Ablation Studies

We performed two groups of additional ablation studies to validate the effectiveness of each proposed module in Diffusion-RSCC. We trained the models on the same LEVIR-CC dataset and evaluated the model performance from two aspects: 1) models with different sub-structures and 2) Stacking Self-attention with different numbers of self-attention blocks. The detailed descriptions are as follows:

#### 1) Models with different sub-structures

Table II shows the results from the first ablation study of our model. We devised different model sub-structures to evaluate the impact of individual module on performance. Specifically, we explored different combinations involving ResNet101, CMF (Cross-Mode Fusion), and SSA (Staking Self-Attention) as the feature extractor, cross-modal data mapping module, and conditional mean value estimation module.

*Effects of the SSA Module:* To verify the effectiveness of the proposed Staking Self-Attention (SSA), we devised two methods: ResNet-CMF and ResNet-CMF-BERTEncoder. ResNet-CMF method discards SSA module while ResNet-CMF-BERTEncoder replaces SSA module with the common language encoding method BERTEncoder [44] in the reverse process of mean value estimation. Table II shows the results from ResNet-CMF, ResNet-CMF-BERTEncoder and ResNet-CMF-SSA. Compared with ResNet-CMF-SSA, ResNet-CMF performed poorly on all metrics, indicating that it was unable to generate natural language correctly and demonstrating the capability of SSA module in change caption generation. The results of ResNet-CMF-BERTEncoder outperformes all other models except for ResNet-CMF-SSA. It shows the capability of the BERTEncoder in mean value estimation. However, our ResNet-CMF-SSA (Diffusion-RSCC) achieves the best performance across all evaluations, underscoring the great effectiveness of the SSA module.

*Effects of the ResNet Module:* ResNet101 (Residual Network) [59] is a convolutional neural network with a depth of 152 layers that democratized the concepts of residual learning and skip connections. It has helped extracting object-level feature in previous change captioning tasks [69]. To further demonstrate the effectiveness of ResNet, we compare the ResNet module with mainstream visual model ViT (Vision Transformer) [70]. As show in Table II, the ResNet-CMF-BERTEncoder significantly outperformed ViT-CMF-BERTEncoder across all metrics, indicating the superior effectiveness of the ResNet module in single-temporal image processing.

*Effects of the CMF Block:* To test the effectiveness and robustness of the proposed Cross-Mode Fusion (CMF), We devised an ImgSelfAtt block, inspired by RSICCformer. In ResNet-ImgSelfAtt-CMF-SSA, the ImgSelfAtt is set for capturing dependencies and relationships within single-temporal image feature $\varphi(I_{bef})$, $\varphi(I_{aft})$. However, Table II shows that the evaluation results of ResNet-ImgSelfAtt-CMF-SSA are significantly lower than those of the ResNet-CMF-SSA. This suggests that the ImgSelfAtt block might lead to overfitting within single-temporal image features, reducing the accuracy of caption generation in bi-temporal scenarios. It also demonstrates that CMF is highly efficient in mapping between bi-temporal image feature $I_{bef}$, $I_{aft}$ and text representation $x_t$.

#### 2) SSA with different numbers of self-attention blocks

For the ablation study on the number of self-attention blocks in SSA, we evaluated the effectiveness of different configurations by setting the number of self-attention blocks from 1 to 5. As shown in Table III, models' performance across all metrics significantly improves as the number of self-attention blocks increases, peaking when there are three blocks. However, when the number of blocks exceeds three, performance begins to decline as additional blocks are added.

As shown in Fig.7, before the number of blocks reaches three, the caption generation model is improved in both grammar learning and accuracy under semantic changes. Beyond three blocks, additional block can enhance the learning of language rules and the flexibility of caption generation, but it negatively impacts the accuracy of caption generation. Through extensive experimentation and careful analysis, we ultimately settled on using the SSA module with three self-attention blocks in our proposed Diffusion-RSCC model.

TABLE II
ABLATION STUDY ON DIFFUSION-RSCC SUB-STRUCTURES. OUR PROPOSED MODEL OUTPERFORMS ALL BASELINES ON ALL THE METRICS. THE NUMBERS ARE IN %. THE BEST RESULTS ARE IN **BOLD**

| Method | BLEU-4 | METEOR | ROUGE-L | CIDEr | SPICE | BARTScore | MoverScore |
|---|---|---|---|---|---|---|---|
| ResNet-CMF | 3.4 | 2.5 | 7.9 | 2.9 | 1.5 | -571.0 | 3.6 |
| ViT-CMF-BERTEncoder | 37.5 | 26.7 | 52.8 | 78.32 | 18.4 | -454.0 | 25.89 |
| ResNet-ImgSelfAtt-CMF-SSA | 37.9 | 29.19 | 56.6 | 87.6 | 20.7 | -444.5 | 27.2 |
| ResNet-CMF-BERTEncoder | 55.3 | 36.9 | 70.1 | 120.4 | 26.6 | -417.7 | 32.2 |
| ResNet-CMF-SSA(Ours) | **60.9** | **37.8** | **71.5** | **125.6** | **26.7** | **-417.7** | **32.4** |

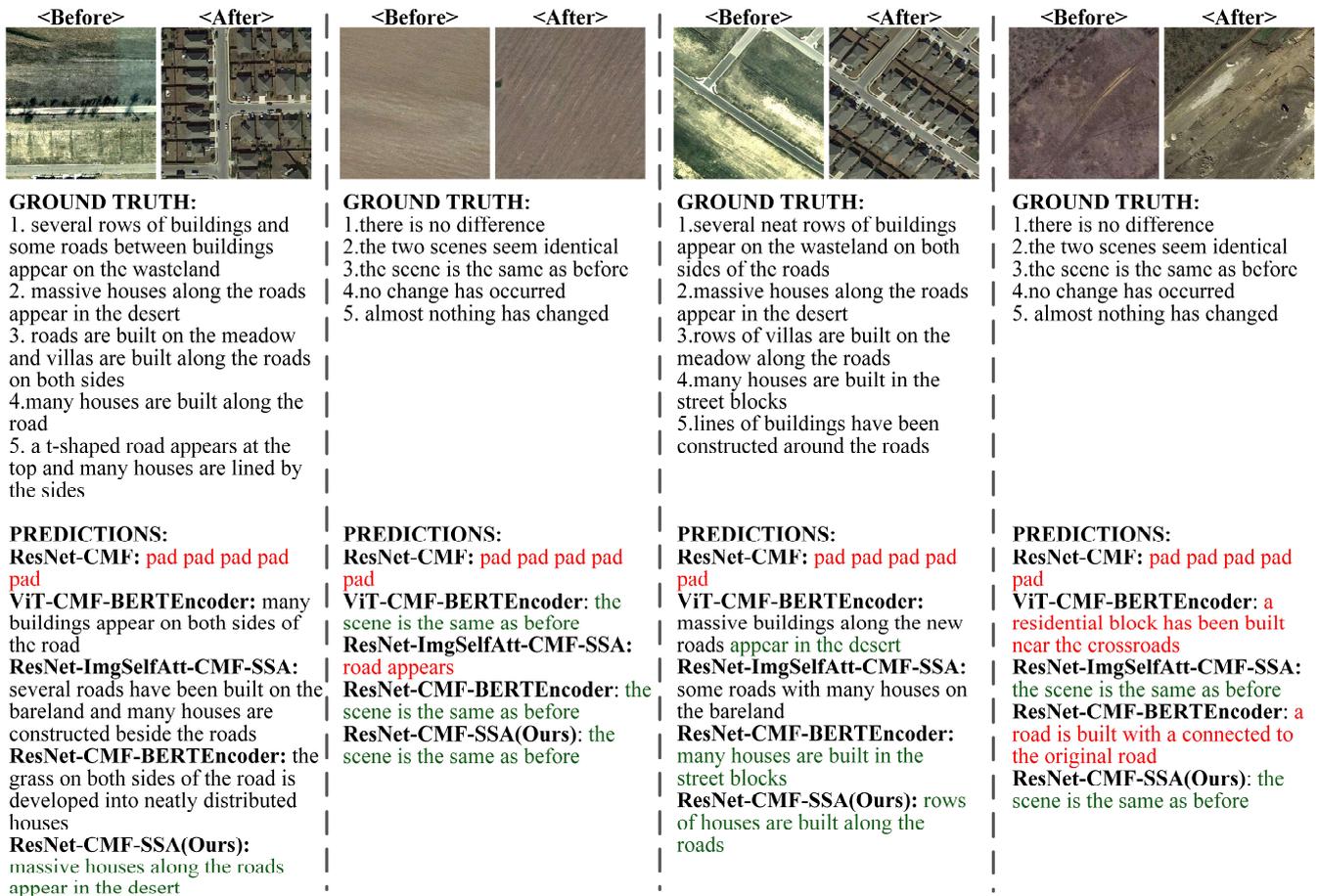

Fig. 6. Examples of the ablation study with different sub-structures. We report the captions of our method and baselines along with the ground truth. Correct sequences are in green and incorrect parts of the captions are in red.

TABLE III
ABLATION STUDY ON STAKING SELF-ATTENTION WITH DIFFERENT NUMBERS OF SELF-ATTENTION BLOCKS. OUR PROPOSED MODEL OUTPERFORMS ALL BASELINES ON ALL THE METRICS. THE NUMBERS ARE IN %. THE BEST RESULTS ARE IN **BOLD**

| Num | BLEU-4 | METEOR | ROUGE-L | CIDEr | SPICE | BARTScore | MoverScore |
|---|---|---|---|---|---|---|---|
| 1 | 50.2 | 29.2 | 56.5 | 96.3 | 21 | -452.1 | 26.1 |
| 2 | 57.4 | 35.7 | 68.2 | 119.7 | 25.7 | -423.7 | 31.1 |
| **3** | **60.9** | **37.8** | **71.5** | **125.6** | **26.7** | **-417.7** | **32.4** |
| 4 | 55.8 | 35.5 | 68 | 117.9 | 25.6 | -424.9 | 31 |
| 5 | 50.9 | 34.7 | 66.9 | 113.1 | 25.4 | -424.6 | 30.5 |

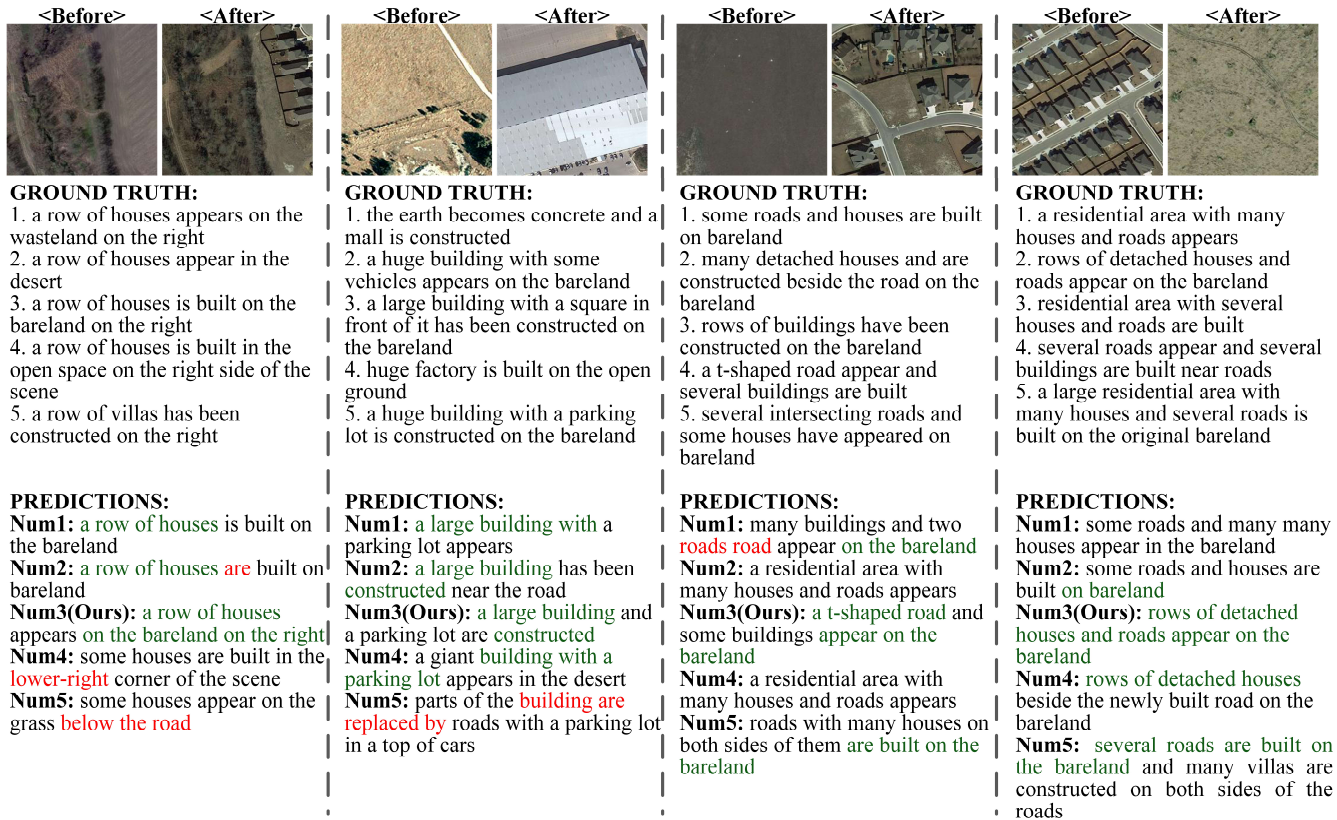

Fig. 7. Examples of the ablation study about SSA with different numbers of self-attention blocks. We report the captions of our method and 1-5 blocks baselines along with the ground truth. Correct sequences are in green and incorrect parts of the captions are in red.

## V. CONCLUSION

In this paper, we propose a novel diffusion model (Diffusion-RSCC) for remote sensing image change captioning to learn the probabilistic distribution of input through forward noising process and reverse denoising process. In the reverse process, a condition denoiser is designed for the mean value estimation conditioned on the bi-temporal image pairs. The image feature extractor ResNet101 module and the newly proposed CMF and SSA in condition denoiser were evaluated on the LEVIR-CC dataset. Through extensive experiments and analyses, we demonstrate the effectiveness of our proposed Diffusion-RSCC and each its individual modules. Diffusion-RSCC achieves superior performance compared to previous change captioning methods.

## REFERENCES

[1] X. Ma, X. Zhang, M. O. Pun, and M. Liu, "A Multilevel Multimodal Fusion Transformer for Remote Sensing Semantic Segmentation," *IEEE Transactions on Geoscience and Remote Sensing,* vol. 62, pp. 1-15, 2024.

[2] A. Zavras, D. Michail, B. Demir, and I. J. a. e.-p. Papoutsis, "Mind the Modality Gap: Towards a Remote Sensing Vision-Language Model via Cross-modal Alignment," 2024, arXiv:2402.09816.

[3] Y. Yang, H. Wei, H. Zhu, D. Yu, H. Xiong, and J. Yang, "Exploiting Cross-Modal Prediction and Relation Consistency for Semisupervised Image Captioning," *IEEE Transactions on Cybernetics,* vol. 54, no. 2, pp. 890-902, 2024.

[4] S. Chang and P. Ghamisi, "Changes to Captions: An Attentive Network for Remote Sensing Change Captioning," *IEEE Transactions on Image Processing,* vol. 32, pp. 6047-6060, 2023.

[5] C. Liu, R. Zhao, J. Chen, Z. Qi, Z. Zou, and Z. Shi, "A Decoupling Paradigm With Prompt Learning for Remote Sensing Image Change Captioning," *IEEE Transactions on Geoscience and Remote Sensing,* vol. 61, pp. 1-18, 2023.

[6] C. Liu, R. Zhao, H. Chen, Z. Zou, and Z. Shi, "Remote Sensing Image Change Captioning With Dual-Branch Transformers: A New Method and a Large Scale Dataset," IEEE Transactions on Geoscience and Remote Sensing, vol. 60, pp. 1-20, 2022.

[7] J. Shi, W. Liu, H. Shan, E. Li, X. Li, and L. Zhang, "Remote Sensing Scene Classification Based on Multibranch Fusion Attention Network," IEEE Geoscience and Remote Sensing Letters, vol. 20, pp. 1-5, 2023.

[8] A. Tahraoui, R. Kheddam, and A. Belhadj-Aissa, "Land Change Detection in Sentinel-2 Images Using IR-MAD And Deep Neural Network," in 2023 International Conference on Earth Observation and Geo-Spatial Information (ICEOGI), 2023, pp. 1-6.

[9] Y. Kim, J. Kim, B. K. Lee, S. Shin, and Y. M. Ro, "Mitigating Dataset Bias in Image Captioning Through Clip Confounder-Free Captioning Network," in 2023 IEEE International Conference on Image Processing (ICIP), 2023, pp. 1720-1724.

[10] S. H. Khan, X. He, F. Porikli, and M. Bennamoun, "Forest Change Detection in Incomplete Satellite Images With Deep Neural Networks," IEEE Transactions on Geoscience and Remote Sensing, vol. 55, no. 9, pp. 5407-5423, 2017.

[11] H. Chen and Z. Shi, "A Spatial-Temporal Attention-Based Method and a New Dataset for Remote Sensing Image Change Detection," Remote Sensing, vol. 12, p. 1662, 05/22 2020.

[12] H. Jhamtani and T. Berg-Kirkpatrick, "Learning to Describe Differences Between Pairs of Similar Images," in Proceedings of the 2018 Conference on Empirical Methods in Natural Language Processing, 2018, pp. 4024–4034.

[13] D. H. Park, T. Darrell, and A. Rohrbach, "Robust Change Captioning," in 2019 IEEE/CVF International Conference on Computer Vision (ICCV), 2019, pp. 4623-4632.

[14] Y. Qiu, Y. Satoh, R. Suzuki, K. Iwata, and H. Kataoka, "3D-Aware Scene Change Captioning From Multiview Images," IEEE Robotics and Automation Letters, vol. 5, no. 3, pp. 4743-4750, 2020.

[15] Y. Qiu et al., "3D Change Localization and Captioning from Dynamic Scans of Indoor Scenes," in 2023 IEEE/CVF Winter Conference on Applications of Computer Vision (WACV), 2023, pp. 1176-1185.

[16] Y. Tu, L. Li, L. Su, J. Du, K. Lu, and Q. Huang, "Viewpoint-Adaptive Representation Disentanglement Network for Change Captioning," IEEE Transactions on Image Processing, vol. 32, pp. 2620-2635, 2023.

[17] Y. Tu et al., "Semantic Relation-aware Difference Representation Learning for Change Captioning," in Findings of the Association for Computational Linguistics: ACL-IJCNLP 2021, 2021, pp. 63–73.

[18] M. Hosseinzadeh and Y. Wang, "Image Change Captioning by Learning from an Auxiliary Task," in 2021 IEEE/CVF Conference on Computer Vision and Pattern Recognition (CVPR), 2021, pp. 2724-2733.

[19] K. E. Ak, Y. Sun, and J. H. Lim, "Learning by Imagination: A Joint Framework for Text-Based Image Manipulation and Change Captioning," IEEE Transactions on Multimedia, vol. 25, pp. 3006-3016, 2023.

[20] S. Chouaf, G. Hoxha, Y. Smara, and F. Melgani, "Captioning Changes in Bi-Temporal Remote Sensing Images," in 2021 IEEE International Geoscience and Remote Sensing Symposium IGARSS, 2021, pp. 2891-2894.

[21] C. Liu, K. Chen, H. Zhang, Z. Qi, Z. Zou, and Z. J. a. e.-p. Shi, "Change-Agent: Towards Interactive Comprehensive Remote Sensing Change Interpretation and Analysis," 2024, arXiv:2403.19646.

[22] C. Liu, K. Chen, H. Zhang, Z. Qi, Z. Zou, and Z. J. a. e.-p. Shi, "Pixel-Level Change Detection Pseudo-Label Learning for Remote Sensing Change Captioning,"2023, arXiv:2312.15311.

[23] W. Peng, P. Jian, Z. Mao, and Y. Zhao, "Change Captioning for Satellite Images Time Series," IEEE Geoscience and Remote Sensing Letters, vol. 21, pp. 1-5, 2024.

[24] J. Ho, A. Jain, and P. Abbeel, "Denoising diffusion probabilistic models," in Advances in Neural Information Processing Systems, 2020, vol. 33, pp. 6840-6851.

[25] S. Gao et al., "Implicit Diffusion Models for Continuous Super-Resolution," in 2023 IEEE/CVF Conference on Computer Vision and Pattern Recognition (CVPR), 2023, pp. 10021-10030.

[26] J. Ho and T. Salimans, "Classifier-Free Diffusion Guidance," in NeurIPS 2021 Workshop on Deep Generative Models and Downstream Applications, 2021.

[27] L. Zheng, J. Yuan, L. Yu, and L. J. a. e.-p. Kong, "A Reparameterized Discrete Diffusion Model for Text Generation," 2023, arXiv:2302.05737.

[28] X. Li, J. Thickstun, I. Gulrajani, P. S. Liang, and T. B. Hashimoto, "Diffusion-LM Improves Controllable Text Generation," in Advances in Neural Information Processing Systems, 2022, vol. 35, pp. 4328–4343.

[29] S. Gong, M. Li, J. Feng, Z. Wu, and L. Kong, "DiffuSeq: Sequence to Sequence Text Generation with Diffusion Models," in International Conference on Learning Representations, ICLR, 2023.

[30] D. Jiang et al., "CoMat: Aligning Text-to-Image Diffusion Model with Image-to-Text Concept Matching," 2024, arXiv:2404.03653.

[31] J. Luo et al., "Semantic-Conditional Diffusion Networks for Image Captioning*," in 2023 IEEE/CVF Conference on Computer Vision and Pattern Recognition (CVPR), 2023, pp. 23359-23368.

[32] G. Hoxha, S. Chouaf, F. Melgani, and Y. Smara, "Change Captioning: A New Paradigm for Multitemporal Remote Sensing Image Analysis," IEEE Transactions on Geoscience and Remote Sensing, vol. 60, pp. 1-14, 2022.

[33] O. Ronneberger, P. Fischer, and T. Brox, "U-Net: Convolutional Networks for Biomedical Image Segmentation," in Medical Image Computing and Computer-Assisted Intervention – MICCAI 2015, 2015, pp. 234-241.

[34] Z. Luo et al., "Notice of Removal: VideoFusion: Decomposed Diffusion Models for High-Quality Video Generation," in 2023 IEEE/CVF Conference on Computer Vision and Pattern Recognition (CVPR), 2023, pp. 10209-10218.

[35] A. Blattmann et al., "Align Your Latents: High-Resolution Video Synthesis with Latent Diffusion Models," in 2023 IEEE/CVF Conference on Computer Vision and Pattern Recognition (CVPR), 2023, pp. 22563-22575.

[36] L. Khachatryan et al., "Text2Video-Zero: Text-to-Image Diffusion Models are Zero-Shot Video Generators," in 2023 IEEE/CVF International Conference on Computer Vision (ICCV), 2023, pp. 15908-15918.

[37] O. Avrahami, D. Lischinski, and O. Fried, "Blended Diffusion for Text-driven Editing of Natural Images," in 2022 IEEE/CVF Conference on Computer Vision and Pattern Recognition (CVPR), 2022, pp. 18187-18197.

[38] N. Tumanyan, M. Geyer, S. Bagon, and T. Dekel, "Plug-and-Play Diffusion Features for Text-Driven Image-to-Image Translation," in 2023 IEEE/CVF Conference on Computer Vision and Pattern Recognition (CVPR), 2023, pp. 1921-1930.

[39] M. Xu, J. Ma, and Y. Zhu, "Dual-Diffusion: Dual Conditional Denoising Diffusion Probabilistic Models for Blind Super-Resolution Reconstruction in RSIs," IEEE Geoscience and Remote Sensing Letters, vol. 20, pp. 1-5, 2023.

[40] C. Saharia, J. Ho, W. Chan, T. Salimans, D. J. Fleet, and M. Norouzi, "Image Super-Resolution via Iterative Refinement," IEEE Transactions on Pattern Analysis and Machine Intelligence, vol. 45, no. 4, pp. 4713-4726, 2023.

[41] M. Wang, Y. Song, P. Wei, X. Xian, Y. Shi, and L. Lin, "IDF-CR: Iterative Diffusion Process for Divide-and-Conquer Cloud Removal in Remote-Sensing Images," IEEE Transactions on Geoscience and Remote Sensing, vol. 62, pp. 1-14, 2024.

[42] Y. Xiao, Q. Yuan, K. Jiang, J. He, X. Jin, and L. Zhang, "EDiffSR: An Efficient Diffusion Probabilistic Model for Remote Sensing Image Super-Resolution," IEEE Transactions on Geoscience and Remote Sensing, vol. 62, pp. 1-14, 2024.

[43] L. Zhang, A. Rao, and M. Agrawala, "Adding Conditional Control to Text-to-Image Diffusion Models," in 2023 IEEE/CVF International Conference on Computer Vision (ICCV), 2023, pp. 3813-3824.

[44] J. Devlin, M.-W. Chang, K. Lee, and K. Toutanova, "BERT: Pre-training of Deep Bidirectional Transformers for Language Understanding," Minneapolis, Minnesota, 2019, pp. 4171-4186.

[45] A. Nichol et al., "GLIDE: Towards Photorealistic Image Generation and Editing with Text-Guided Diffusion Models," 2021, arXiv.2112.10741

[46] S. Saha, F. Bovolo, and L. Bruzzone, "Unsupervised Deep Change Vector Analysis for Multiple-Change Detection in VHR Images," IEEE Transactions on Geoscience and Remote Sensing, vol. 57, no. 6, pp. 3677-3693, 2019.

[47] C. Wu, H. Chen, B. Du, and L. Zhang, "Unsupervised Change Detection in Multitemporal VHR Images Based on Deep Kernel PCA Convolutional Mapping Network," IEEE Transactions on Cybernetics, vol. 52, no. 11, pp. 12084-12098, 2022.

[48] B. Du, Y. Wang, C. Wu, and L. Zhang, "Unsupervised Scene Change Detection via Latent Dirichlet Allocation and Multivariate Alteration Detection," IEEE Journal of Selected Topics in Applied Earth Observations and Remote Sensing, vol. 11, no. 12, pp. 4676-4689, 2018.

[49] Y. Zeng, M. E. Schaepman, B. Wu, J. G. P. W. Clevers, and A. K. Bregt, "Scaling-based forest structural change detection using an inverted geometric-optical model in the Three Gorges region of China," Remote Sensing of Environment, vol. 112, no. 12, pp. 4261-4271, 2008.

[50] X. Peng, R. Zhong, Z. Li, and Q. Li, "Optical Remote Sensing Image Change Detection Based on Attention Mechanism and Image Difference," IEEE Transactions on Geoscience and Remote Sensing, vol. 59, no. 9, pp. 7296-7307, 2021.

[51] M. Liu, Q. Shi, Z. Chai, and J. Li, "PA-Former: Learning Prior-Aware Transformer for Remote Sensing Building Change Detection," IEEE Geoscience and Remote Sensing Letters, vol. 19, pp. 1-5, 2022.

[52] R. C. Daudt, B. L. Saux, A. Boulch, and Y. Gousseau, "Guided Anisotropic Diffusion and Iterative Learning for Weakly Supervised Change Detection," in 2019 IEEE/CVF Conference on Computer Vision and Pattern Recognition Workshops (CVPRW), 2019, pp. 1461-1470.

[53] J. Tian, J. Lei, J. Zhang, W. Xie, and Y. Li, "SwiMDiff: Scene-Wide Matching Contrastive Learning With Diffusion Constraint for Remote Sensing Image," IEEE Transactions on Geoscience and Remote Sensing, vol. 62, pp. 1-13, 2024.

[54] X. Shi, X. Yang, J. Gu, S. Joty, and J. Cai, "Finding It at Another Side: A Viewpoint-Adapted Matching Encoder for Change Captioning," in Computer Vision – ECCV 2020, 2020, pp. 574-590.

[55] H. Kim, J. Kim, H. Lee, H. Park, and G. Kim, "Viewpoint-Agnostic Change Captioning with Cycle Consistency," in 2021 IEEE/CVF International Conference on Computer Vision (ICCV), 2021, pp. 2075-2084.

[56] Y. Tu, L. Li, C. Yan, S. Gao, and Z. Yu, "R$^3$Net:Relation-embedded Representation Reconstruction Network for Change Captioning," in Proceedings of the 2021 Conference on Empirical Methods in Natural Language Processing, 2021, pp. 9319–9329.

[57] Y. Qiu et al., "Describing and Localizing Multiple Changes with Transformers," in 2021 IEEE/CVF International Conference on Computer Vision (ICCV), 2021, pp. 1951-1960.

[58] C. Luo, Understanding Diffusion Models: A Unified Perspective. 2022, arXiv:2208.11970.

[59] K. He, X. Zhang, S. Ren, and J. Sun, "Deep Residual Learning for Image Recognition," in 2016 IEEE/CVF Conference on Computer Vision and Pattern Recognition (CVPR), 2016, pp. 770-778.
[60] A. Vaswani et al., "Attention is all you need," in Proceedings of the 31st International Conference on Neural Information Processing Systems, Long Beach, California, USA, 2017, pp. 6000–6010.
[61] V. Meshchaninov et al., "Diffusion on language model embeddings for protein sequence generation," 2024, arXiv:2403.03726.
[62] S. Yue, Y. Tu, L. Li, Y. Yang, S. Gao, and Z. Yu, "I3N: Intra- and Inter-Representation Interaction Network for Change Captioning," IEEE Transactions on Multimedia, vol. 25, pp. 8828-8841, 2023.
[63] K. Papineni, S. Roukos, T. Ward, and W.-J. Zhu, "Bleu: a Method for Automatic Evaluation of Machine Translation," in Proceedings of the 40th Annual Meeting of the Association for Computational Linguistics, 2002, pp. 311–318.
[64] C.-Y. Lin, "ROUGE: A Package for Automatic Evaluation of Summaries," in Text Summarization Branches Out, 2004, pp. 74–81.
[65] S. Banerjee and A. Lavie, "METEOR: An Automatic Metric for MT Evaluation with Improved Correlation with Human Judgments," in Proceedings of the ACL Workshop on Intrinsic and Extrinsic Evaluation Measures for Machine Translation and/or Summarization, 2005, pp. 65–72.
[66] R. Vedantam, C. L. Zitnick, and D. Parikh, "CIDEr: Consensus-based image description evaluation," in 2015 IEEE Conference on Computer Vision and Pattern Recognition (CVPR), 2015, pp. 4566-4575.
[67] C. Niu, H. Shan, and G. Wang, "SPICE: Semantic Pseudo-Labeling for Image Clustering," IEEE Transactions on Image Processing, vol. 31, pp. 7264-7278, 2022.
[68] W. Yuan, G. Neubig, and P. Liu, "BARTScore: Evaluating Generated Text as Text Generation," in Advances in Neural Information Processing Systems, vol. 34, pp. 27263–2727, 2021.
[69] Y. Tu, L. Li, L. Su, Z. J. Zha, C. Yan, and Q. Huang, "Self-supervised Cross-view Representation Reconstruction for Change Captioning," in 2023 IEEE/CVF International Conference on Computer Vision (ICCV), 2023, pp. 2793-2803.
[70] A. Berroukham, K. Housni, and M. Lahraichi, "Vision Transformers: A Review of Architecture, Applications, and Future Directions," in 2023 7th IEEE Congress on Information Science and Technology (CiSt), 2023, pp. 205-210.

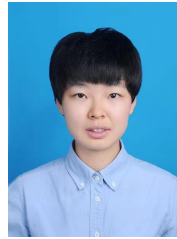

**Jie Ma** (Member, IEEE) received the B.S. degree in mathematics and applied mathematics from Capital Normal University, Beijing, China, in 2015, and the Ph.D. degree in computer application technology from Beijing Normal University, Beijing, China, in 2020.

She is currently an Associate Professor with the School of Information Science and Technology, Beijing Foreign Studies University, Beijing, China. She has authored or coauthored over 10 scientific papers in refereed journals and proceedings, including the IEEE TRANSACTIONS ON GEOSCIENCE AND REMOTE SENSING (TGRS), IEEE GEOSCIENCE AND REMOTE SENSING LETTERS (GRSL). Her research interests include computer vision, multi-modal tasks and related problems in remote sensing.

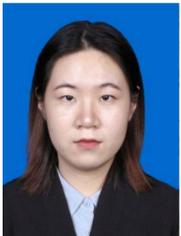

**Xiaofei Yu** received the B.S. degree from the School of Information Management and Engineering, Shanghai University of Finance and Economics, Shanghai, China, in 2022. She is currently pursuing the M.S. degree at the School of Information Science and Technology, Beijing Foreign Studies University, Beijing, China.

Her research interests include computer vision, deep learning, and related problems in remote sensing remote sensing multi-modal tasks.

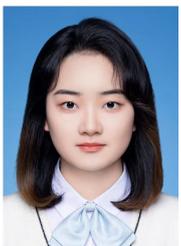

**Yitong Li** received the B.S. degree from the School of Information Science and Technology, Beijing Foreign Studies University, Beijing, China, in 2024. She is currently pursuing the M.S. degree at the State Key Laboratory of Information Engineering in Surveying, Mapping and Remote Sensing, Wuhan University, Wuhan, China.

Her research interests include computer vision, deep learning, and related problems in remote sensing.